# Urban Deceleration Behavior Modes Under Scene Context: An Early-Kinematic Classifier from Argoverse 2 Multi-Agent Trajectories

**Eni Solomon Laughter 1,***

1 School of Transportation Engineering, Chang'an University, Xi'an, Shaanxi, China; *2024134912@chd.edu.cn*
* Correspondence: *2024134912@chd.edu.cn*;

**Abstract:** Urban deceleration is one of the most empirically studied yet least taxonomically organized behaviors in car-following research. With recent perception-equipped autonomous-vehicle datasets, enable trajectory-anchored mode discovery. We extract 1,219 sustained deceleration events from 234 urban driving logs of the Argoverse 2 Sensor dataset, encode each event in a 19-dimensional kinematic feature vector, discover behavioural modes via K-means clustering with bootstrap stability analysis, and quantify modulation by eleven scene-context variables. A HistGradientBoosting classifier predicts mode membership from the first 1.0 s of each event. Four stable modes emerge with bootstrap Adjusted Rand Index of 0.897 across 50 resamples — anticipatory soft (62.8%), reactive closing (30.6%), brake-like jerk (4.8%), and an outlier category (1.8%). Only pair age shows a medium effect ($\varepsilon^2 = 0.085$); scene geometry and vulnerable-road-user proximity show negligible effects. The early-event classifier achieves macro-F1 = 0.758 at 1.0 s, with scene context contributing +0.059 F1 over kinematics alone. Modes are regime-invariant in medium-speed driving (ARI = 0.817) but regime-dependent at low speed (ARI = 0.166). A small set of stable kinematic modes structures urban deceleration; early-window jerk dominates predictive signal; and pair age is the primary contextual modulator.

## 1. Introduction

What drivers do with their longitudinal control during when they drive, when they brake, how sharply, and in response to what visual and contextual cues, is one of the clearest behavioral signals captured by vehicle trajectory dataset. This control kinematic record is preserved at high temporal resolution, and the rapid maturation of perception-equipped autonomous-vehicle datasets has shifted the empirical frontier from simulator-based and instrumented-vehicle studies toward ground-truth-annotated multi-agent trajectory data. Where simulator studies offer controllability of stimuli and counterfactual replay, and where instrumented-vehicle naturalistic studies offer pedal-level fidelity, neither operates at the scale or scene diversity afforded by perception-AV datasets, in which dozens of co-present agents are tracked simultaneously with object class, geometry, and high-definition map context.

Trajectory data drawn from such datasets preserve the actual behavioral choices made by real drivers in real urban scenes, with measured rather than scripted contextual conditions, and without the behavioral changes that can occur when a driver knows they are operating an instrumented vehicle. Reaction-time microstructure, looming-driven brake initiation, and stop-phase transitions — all characterized under controlled or single-vehicle naturalistic conditions [1–3] manifest in trajectory data as observable kinematic signatures of behavioral modes. These properties motivate the present study's use of the Argoverse 2 Sensor dataset [4] rather than simulator-generated or single-vehicle naturalistic data.

The remainder of this introduction first lays out the conceptual background that underpins this study (Section 1.1) and then states the research objective and contributions to the growing body of trajectory-driving-behaviour work (Section 1.2).

### *1.1. Background*

Car-following modelling has progressed through fifty years of architectural refinement. From the constant-time-headway formulations of the 1950s, through the stimulus–response Gazis–Herman–Rothery family, the optimum-velocity formulation of Bando and colleagues, the Wiedemann threshold-based psycho-physical model, and into the Intelligent Driver Model that became the modern empirical baseline. This modeling is supported with the arrival of large-scale trajectory data with NGSIM in the mid-2000s which shifted the methodological premise from analytic derivation to data-driven calibration, and the refinements that followed have each introduced a specific mechanism onto this base rather than replacing it wholesale. Li et al. [5] embedded a space-gap-influenced response function into the velocity–spacing–acceleration coupling, demonstrating that follower deceleration realism under variable headways improves when the response itself is allowed to vary with gap rather than holding fixed across the spacing range. Hamdar et al. [6] subsequently allowed environmental conditioning such as weather and road geometry — to enter the longitudinal acceleration framework as covariates, evidencing that ambient conditions modulate the stimulus–response coupling. Chen et al. [7] consolidated these threads through the FollowNet benchmark, which assembled five trajectory datasets under a unified evaluation protocol and standardized how modern car-following models are calibrated and tested against naturalistic data. The current frontier is explicitly multi-vehicle: Das et al. [8] introduced a two-leader recurrent network that encodes the positional arrangement of multiple leaders into the follower's state, finding that two-leader information improves prediction beyond the conventional single-leader assumption; Lyu et al. [9] built driver heterogeneity directly into the prediction architecture by coupling vehicle-dynamics features with contrastive-learning-derived preference embeddings, rather than treating heterogeneity as residual noise; and Chen et al. [10] synthesized this trajectory in a recent review, identifying multi-vehicle interaction under heterogeneous traffic as the principal direction in which the field is now moving.

Within the deceleration subdomain, empirical work has organized itself around two complementary axes: the temporal *profile* of the event (how the magnitude unfolds in time) and the underlying *mechanism* (what physical action produces the kinematic signature). The profile perspective was concretized by Deligianni et al. [11], who used naturalistic trajectory observations to identify onset, peak, and recovery as the three principal axes of profile variation in normal driving, providing the methodological template that subsequent work extended. Liu and Zhang [12] adapted this template to high-speed environments and characterized the distributional shift of profile parameters across speed regimes; Li et al. [13] later formalized the temporal organization of pre-collision deceleration sequences via random-parameters survival modelling on NGSIM, demonstrating that transition durations between profile phases themselves carry predictive information about rear-end risk. The mechanism perspective developed in parallel. Da Lio et al. [14] proposed a biologically guided decomposition of stop behaviour into a small set of discrete stop phases, arguing that drivers transition through distinct control regimes rather than producing a continuous response. Li et al. [15] supported this view empirically by partitioning naturalistic deceleration events into no-pedal-input (coasting) and brake-pedal-input categories using pedal-instrumented vehicles, demonstrating that the kinematic signature of the mechanism is concentrated in the jerk profile: coasting produces gradual jerk, while brake application produces sharper jerk peaks. Subsequent contributions grounded these conceptual tools in safety-relevant urban settings — Wang et al. [16] coupled daily car-following behaviour to rear-end crash and near-crash outcomes; Yang et al. [17] applied LiDAR-based trajectory extraction to urban intersection deceleration; Yamagishi [18] characterized rapid-deceleration events with attention to driver age cohorts. Together these contributions established that urban deceleration is jointly characterized by its profile shape and its mechanism — an insight the present study captures through the Family B feature design.

How drivers perceive the need to decelerate has been investigated through a methodological progression that increasingly localizes the cognitive stimulus in measurable visual quantities. Xue et al. [1] characterized the role of visual looming — "the retinal rate of expansion of the lead vehicle" — as the primary kinematic stimulus for rear-end collision avoidance, demonstrating through combined simulator and trajectory analysis that looming explains the timing of brake initiation more reliably than relative velocity alone. Durrani et al. [2] then questioned the dominant fixed-threshold reaction-time paradigm, formulating perception–reaction as an evidence-accumulation process and showing that such models substantially outperform threshold formulations in predicting actual brake-onset timing across multiple datasets. Svärd et al. [3] extended the analysis to peripheral vision through controlled glance-response experiments, finding that critical-deceleration responses can be initiated from peripheral visual cues alone, with glance-response times largely independent of visual eccentricity — implying that the early portion of a deceleration event is informative even when foveal attention is elsewhere. The most recent extension comes from Khakzar et al. [19], who applied a deep-learning architecture combining LSTM trajectory encoding with a convolutional risk-feature input to naturalistic car-following sequences, identifying driver's behaviour as a substantive moderator of trajectory-prediction accuracy and demonstrating that modern architecture can integrate the looming signal with broader contextual features. Together these contributions establish the methodological premise underlying early-prediction approaches: the kinematic state during the first second of a deceleration event carries substantial information about the response that follows, and this information is recoverable from trajectory data without recourse to gaze or pedal sensors.

Driving behaviour has been organized through unsupervised methods for over a decade, and the methodological trajectory of this body of work has progressed from coarse style partitioning toward context-aware behavioral taxonomies. The foundational work of Chen and Chen [20] applied clustering to naturalistic instrumented-vehicle data and identified persistent style modes — establishing both the small-cardinality finding and the methodological template that subsequent studies extended. Wang et al. [21] adapted this template to connected-vehicle driving-encounter scenarios, demonstrating that the clustering paradigm transfers from single-driver style characterization to multi-agent interaction patterns. Methodological refinements followed: de Zepeda et al. [22] introduced a dynamic clustering framework in which mode membership evolves over time rather than being fixed per driver, while Ali et al. [23] developed risky-versus-normal time-series clustering on freeway segments, evidencing that weather and roadway conditions partition the kinematic signal. Subsequent work conditioned these methods on specific contexts. Khanfar et al. [24] applied unsupervised methods to signalized-intersection deceleration; Lee and Jang [25] characterized longitudinal naturalistic behaviour with an autoencoder-followed-by-HDBSCAN architecture; Pan et al. [26] coupled driving-style clustering to conflict-risk prediction with driving style as a covariate; Cui et al. [27] introduced time-pressure-induced aggression as a context dimension and used SHAP attribution to interpret the resulting classifier; Chen et al. [28] extended the partitioning into the safety-evaluation domain through trajectory-anchored risk clustering. A complementary methodological development emerged in urban traffic-state clustering: Wang and Cheng [29] incorporated lane-changing-behaviour indicators alongside conventional flow-density-occupancy variables in a B-K-means partition, demonstrating that behavioral-state features sharpen the discrimination between free-flow, synchronized-flow, and wide-moving-jam regimes beyond what aggregate flow variables alone support. A supervised-side methodological parallel comes from Yu et al. [30], who used random-forest classification on naturalistic-driving inputs to predict speeding behaviour, evidencing that tree-ensemble classifiers transfer cleanly from environmental and contextual features to behavioral targets — the methodological premise on which the present study's early-event classifier rests.

Within the closely connected lane-change subdomain to which a portion of the deceleration literature is linked through the cut-in pathway, three methodological lines have developed in parallel. Guo et al. [31] reformulated lane-change detection as a one-class anomaly-detection problem solved by autoencoders, treating routine lane-keeping as the in-distribution baseline against which maneuvers stand out; Gu et al. [32] used random forests on NGSIM trajectories to predict lane-change decisions from kinematic and contextual inputs, demonstrating the supervised counterpart; Hill et al. [33] documented driver-type-stratified heterogeneity in lane-change kinematics on freeway data, evidencing that the modes uncovered in clustering studies have a real behavioral substrate. The empirical heterogeneity analysis of Ossen and Hoogendoorn [34] — based on a large helicopter-collected trajectory sample and joint calibration of eight different car-following formulations — established the strongest finding in this body of work: that passenger-car drivers differ not only in parameter values within a single model architecture, but in the very stimuli to which they respond, with different drivers selecting different lead-vehicle cues as the dominant input to their longitudinal control. This supports the view that behavioral-mode boundaries reflect inherent driving-style heterogeneity rather than merely parameter dispersion. The systematic review by Bouhsissin et al. [35] consolidates this entire body of work and confirms that a three-to-four-mode cardinality recurs across data sources and methodologies, commonly interpreted along axes such as aggressive–normal–cautious or proactive–reactive.

The scene surrounding each deceleration event has been progressively articulated as a measurable modulator of follower kinematics rather than treated as residual variation, with successive contributions adding specific context dimensions to the predictive vocabulary. Fu et al. [36] developed a human-like car-following model that explicitly incorporated lead-vehicle cut-in behaviour into the follower's stimulus, demonstrating that cut-in-induced decelerations differ systematically from continuous-following decelerations in both intensity and timing. Hu et al. [37] generalized this to predictive models of follower acceleration and deceleration responses across both cut-in and cut-out maneuvers of the lead vehicle, evidencing that the symmetric treatment of insertion and exit is necessary for accurate response prediction. The lane-change-stimulus pathway was further refined by Zhao et al. [38], whose microscopic-trajectory analysis formalized the stimulus the following vehicle receives during a preceding lane change, showing that follower acceleration responses depend on the type of stimulus and the space-gain effect produced by the lane-passing vehicle. Gu et al. [39] extended this to consecutive lane-changing scenarios on HighD data, finding that the stimulative effect on following drivers is heterogeneous and shaped by psychological factors that vary across drivers. Jokhio et al. [40] examined the role of surrounding traffic in lane-change decisions through a video-based laboratory study, finding that lag-vehicle behaviour exerts greater influence on the decision than lead-vehicle behaviour and that lag-vehicle deceleration to create a gap is perceived by other drivers as cooperative. Methodologically adjacent contributions include the lane-change-decision categorization of Keyvan-Ekbatani et al. [41], which documented systematic driver-strategy heterogeneity in lane-change decision-making, and the work of Mullakkal-Babu et al. [42], who characterized fragmented lane changes — maneuvers in which the lane change is initiated then interrupted — as kinematically distinct from continuous lane changes. A parallel line developed around the pedestrian–vehicle interface. Noonan et al. [43] established empirically that drivers communicate yielding intent through deceleration kinematics, with onset timing and magnitude functioning as signals to crossing pedestrians; Chen et al. [44] characterized how deceleration onset timing and approach velocity shape pedestrian perceived safety and comfort during autonomous-vehicle yielding scenarios. Qin et al. [45] contributed a methodological constraint that applies across all of the above: a consistency analysis showing that even within a single driver, car-following behaviour varies systematically across speed regimes — evidencing that any analysis of context modulation must control for the speed regime in which the events occur.

A complementary methodological progression has moved in the direction of *interpretable* behavioral prediction: from black-box risk models toward attribution-anchored taxonomies in which each prediction can be traced back to its driving features. Shi et al. [46] applied XGBoost to driving assessment and risk prediction, establishing that gradient-boosted decision trees can carry the predictive load while remaining structurally interpretable through feature-importance ranking. The introduction of SHAP attribution by Cheng et al. [47] in the context of heavy-vehicle aggressive-driving recognition opened the inner workings of artificial-neural-network classifiers to per-prediction inspection, allowing feature contributions to be audited at the event level; Cui et al. [27] applied the same attribution framework to time-pressure-induced aggression, evidencing the transferability of attribution-based interpretation across behavioral contexts. Lu et al. [48] developed a learned representation of surrogate safety measures that allows traffic-conflict identification to be performed jointly with the SSM feature extraction itself, rather than treating SSMs as fixed analytic constructs imported from earlier conflict-analysis traditions. Within the autonomous-vehicle integration domain, Ali [49] examined the immediate behavioral impact of autonomous-vehicle lane changes on following human drivers, and Ma et al. [50] integrated driving-behaviour pattern recognition directly into arterial trajectory reconstruction, demonstrating that behavioral-mode awareness improves spatial reconstruction itself. Adjacent risk-assessment methodologies have developed a field-theoretic framework that quantifies risk through field primitives rather than discrete event metrics: Joo et al. [51] formulated a generalized driving risk assessment using field theory on high-speed highways; Liu and Xiang [52] extended this to lane-change risk through the F-F diagram, a two-dimensional surrogate safety indicator with explicit field-theory primitives; Wu et al. [53] applied the same primitives to longitudinal car-following risk in autonomous-vehicle settings. Xue et al. [54] proposed an improved lane-change risk estimation model based on naturalistic vehicle trajectories from the HighD dataset, refining the parameter-estimation step that field-theoretic approaches require. Kar et al. [55] applied extreme-value analysis to crash-risk modelling using maximum deceleration rate as a covariate — a methodological choice that directly motivates the inclusion of peak deceleration in the present Family A feature design. Together these contributions evidence that behavioral-mode inference is most usefully coupled with surrogate-safety quantification on one side and attribution-based interpretability on the other.

A parallel evolution in trajectory-data infrastructure has reshaped what behavioral-mode analyses can be empirically grounded in. Li et al. [56] revisited fifteen years of trajectory-data-based traffic-flow research, documenting the field's transition from loop detectors and probe vehicles to NGSIM in the mid-2000s and onward to drone- and perception-AV-collected trajectories at the current frontier — each step bringing finer temporal resolution, broader spatial coverage, and richer per-agent annotation. Berghaus et al. [57] contributed a drone-extracted highway trajectory dataset that includes off-ramp and congested-traffic scenarios with high-definition geometry, evidencing the operational maturity of drone-based trajectory collection for safety-critical scenario coverage. Ahmed et al. [58] synthesized lessons from major naturalistic-driving-study datasets worldwide, identifying systematic differences in dataset design choices — vehicle instrumentation, sampling frequency, contextual annotation depth — and the consequent transferability constraints these differences impose on derived findings. Bharilya et al. [59] applied a self-supervised transformer to the Argoverse 2 trajectory-prediction task with noise-imputed past trajectories, demonstrating that recent perception-AV datasets such as Argoverse 2 now support the same model architectures developed on earlier-generation trajectory data. Rowan et al. [60] surveyed the rise of machine-learning-based microscopic traffic-flow modelling and articulated the convergence between data-driven and physics-based formulations that this generation of datasets has enabled. Together these infrastructure advances establish the modern empirical conditions under which trajectory-based behavioral analyses can be grounded.

Across these methodological threads, the literature has assembled the conceptual and analytical building blocks for a unified question: can urban deceleration be characterized by a small set of stable kinematic modes whose taxonomy is recoverable from trajectory data, whose contextual modulation can be quantified, and whose mode membership is predictable from the first second of observation? Each block already exists in the literature in fully developed form — looming as the visual stimulus that initiates braking, coasting-versus-braking as the mechanism distinction, profile shape and asymmetry as the temporal descriptors, pair age as a cut-in proxy, K-means clustering with bootstrap stability as the partitioning method, SHAP and permutation importance as the interpretability layer, and perception-AV trajectory datasets as the modern empirical substrate. What has not been done is to assemble these specific blocks into a single coherent pipeline and apply that pipeline to urban car-following deceleration on a perception-AV dataset. The present study performs that assembly.

*1.2. Research Objective and Contributions*

This study aims to develop and evaluate a kinematics-first taxonomy of urban car-following deceleration events, characterize the modulation of the discovered modes by structured scene-context variables, and quantify the early-prediction accuracy of mode membership from short observation windows. The methodology assembles established conceptual and analytical tools — each drawn from the threads reviewed above — into a single coherent pipeline applied to a problem setting in which this exact assembly has not previously been used.

Specifically, the present study takes the visual-looming concept of Xue et al. [11] and embeds it as a Family A surrogate-safety feature alongside DRAC and MTTC, following the surrogate-safety-measure conventions of Lu et al. [61]; it takes the coasting-versus-braking mechanism distinction of Li et al. [15] and Da Lio et al. [14] and reformulates it as a pedal-free mechanism proxy $m = \frac{|j_{peak}|}{|a_{peak}|}$ computable from trajectory data alone, capturing the discrete-stop-phase insight without requiring instrumented vehicles; it takes the cut-in-versus-continuous-following distinction of Fu et al. [36] and Hu et al. [37] and reformulates it as a continuous-valued pair-age variable, allowing the cut-in effect further documented by Zhao et al. [38] and Gu et al. [39] to be quantified statistically rather than treated as a binary categorical; it takes the small-cardinality clustering tradition of Chen and Chen [20] and applies K-means with bootstrap Adjusted Rand Index stability analysis at $K \in \{3, 4, 5, 6\}$ alongside HDBSCAN and Gaussian Mixture Model alternatives, following the methodological discipline established by the cluster-stability literature [22,23,35]; it takes the speed-regime stratification protocol of Qin et al. [62] and applies it as the low-, medium-, and high-speed partition for testing regime-dependence of the discovered modes; it takes the attribution-based interpretability framework of Cheng et al. [47] and Cui et al. [27] in its permutation-importance form, applied to a HistGradientBoosting classifier trained on early-window kinematic and context features; and it takes the perception-AV trajectory data of the Argoverse 2 ecosystem [59] as the empirical substrate, providing ground-truth cuboid annotations, multi-agent coverage, and high-definition map context that earlier NGSIM-based deceleration analyses — including the present author's prior highway work [63] — did not have access to.

The assembly of these tools into a single coherent pipeline and their application to 1,219 sustained urban deceleration events drawn from 234 logs of the Argoverse 2 Sensor dataset is the substantive contribution of this study. To the present author's knowledge, this specific combination has not been assembled in this form in prior work.

The study addresses four research questions. RQ1: what stable kinematic modes exist in the urban deceleration data? RQ2: to what extent does scene context modulate the discovered modes? RQ3: can mode membership be predicted from the early portion of a deceleration event, and at what horizon? RQ4: are the discovered modes regime-invariant across speed bins or regime-dependent?

The study delivers three artefacts: the eight-phase processing pipeline reproducible from raw cuboid annotations; the literature-anchored kinematic-and-context feature framework transferable to other trajectory datasets; and the horizon-stratified early-mode classifier with marginal-context interpretation. Section 2 details the dataset, variables, event-selection procedure, and analytical methods. Section 3 reports results for the four research questions. Section 4 discusses the findings, recommendations for future research and practice, and study limitations. Section 5 concludes.

## 2. Materials and Methods

### *2.1. Dataset and Variables*

Empirical analyses are conducted on a 300-log subset of the Argoverse 2 Sensor dataset training split. Each log comprises approximately 15.6 seconds of urban driving captured in six United States cities, with vehicle and vulnerable-road-user cuboid annotations sampled at 10 Hz, high-definition map vector layers (lane segments, pedestrian crossings, intersection geometry), and per-frame ego-vehicle pose in city-frame coordinates. **Figure 1** shows an illustrative scene from the dataset: a bird's-eye-view LiDAR point cloud with ground-truth cuboid annotations for vehicles, pedestrians, and other agents (left panel), and a synchronized forward-facing camera view of the same urban intersection with class-labelled bounding-box overlays (right panel). The Argoverse 2 Sensor dataset is selected for three properties: ground-truth perception (cuboids are derived from manually verified annotations), multi-agent coverage (typically over one hundred simultaneously tracked agents per log), and ground-anchored high-definition map context. LiDAR and camera data are retained off-disk; the present analysis uses only cuboid annotations, ego pose, high-definition map vectors, and calibration. After trajectory reconstruction, the dataset contains 3,605,320 valid agent-frame observations across 37,134 retained vehicle tracks, with a mean of 123.8 retained tracks per log [4,64].

Each detected deceleration event is encoded by a vector of variables organized into three families.

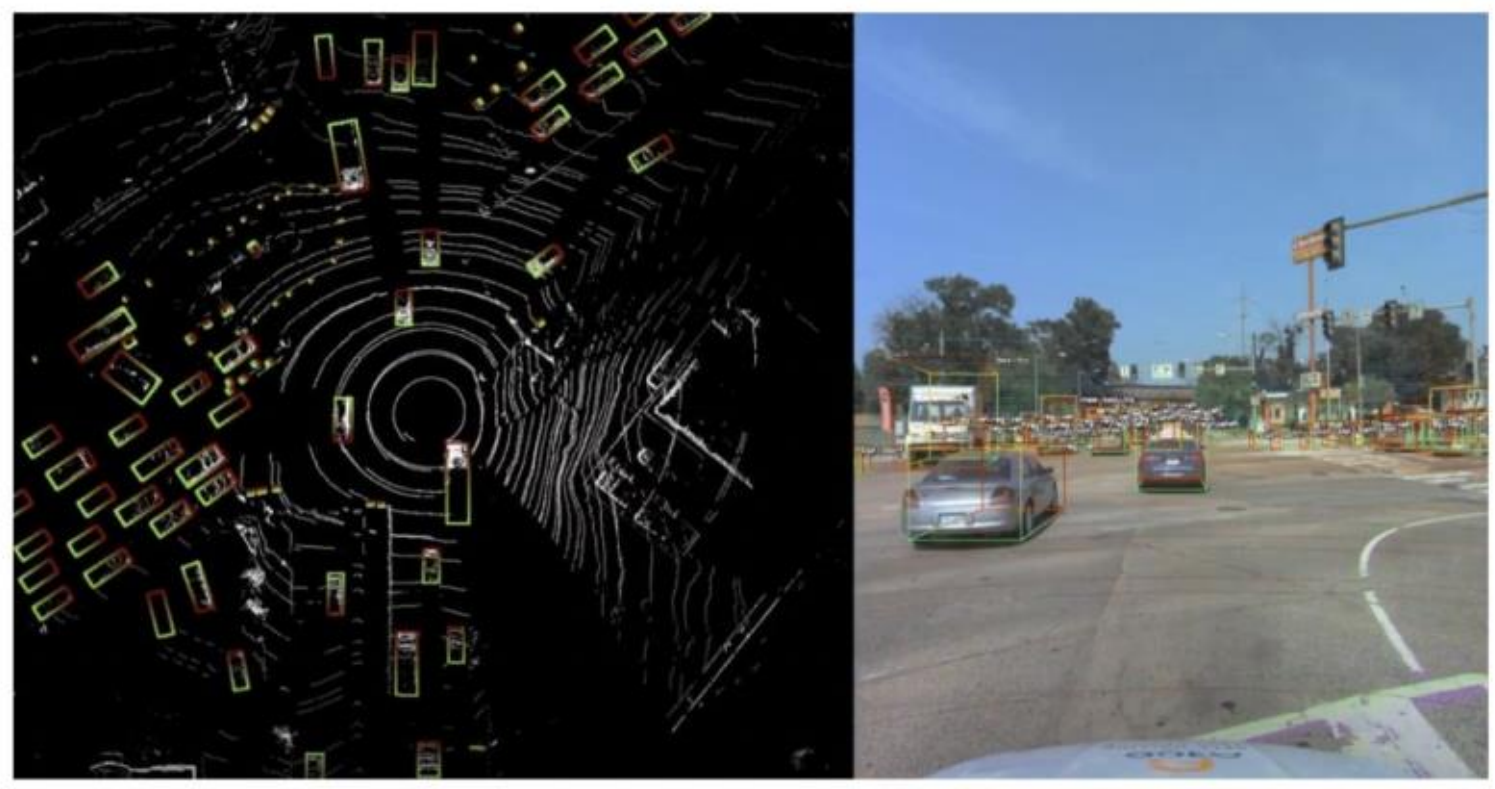

**Figure 1.** Illustrative view of the Argoverse 2 Sensor dataset annotation structure. *Left:* bird's-eye-view BEV LiDAR point cloud with ground-truth 3D cuboid annotations for vehicles, pedestrians, and other agents (green/red boxes). *Right:* synchronized front-facing camera frame at an urban intersection with class-labelled bounding-box overlays. The cuboid annotations provide the per-frame multi-agent trajectory ground truth used in this study; high-definition map vector layers (lane segments, crosswalks, intersection geometry) are provided as a separate vector layer that is not visualized in this view. The raw LiDAR and camera modalities themselves are not used in the present analysis. Image credit goes to the Argo team: Argoverse 2 Sensor dataset [4,64].

#### 2.1.1. Family A — Kinematic Intensity and Timing

Family A captures the magnitude and temporal characteristics of the deceleration — that is, how hard the vehicle braked, how long the event lasted, and how safety-critical the situation looked at the moment it began. The variables used, comprises of peak deceleration $a_{peak}$ and mean deceleration $a_{mean}$ (m/s$^2$); peak jerk $j_{peak}$ and root-mean-square jerk $j_{rms}$ (m/s$^3$); the fraction of the event spent at or below the −0.5 m/s$^2$ detection threshold; event duration (s); the velocity at event onset and the change in velocity over the event (m/s); the initial gap to the leader (m); the initial time-headway (s); the initial gap-closing rate (m/s); the initial visual looming defined in Equation (1); and the deceleration rate to avoid collision (DRAC) defined in Equation (2) along with the modified time-to-collision (MTTC), both following surrogate-safety-measure conventions [61].

$$\lambda = \frac{\dot{g}}{g} \tag{1}$$

where $\lambda$ is the looming rate (s$^{-1}$), g is the longitudinal gap to the leader (m), and $\dot{g}$ is the gap-closing rate (m/s).

$$DRAC = \frac{(\Delta v)^2}{2g}, when\ \Delta v > 0 \tag{2}$$

where Δv is the gap-closing rate (positive when the follower is approaching the leader). DRAC carries units of m/s$^2$ and represents the constant deceleration the follower would need to apply to avoid collision under the assumption that the leader maintains its current speed. MTTC extends the time-to-collision formulation by accounting for relative acceleration; it is the smallest positive root of $g + \Delta v \cdot t + \frac{1}{2}\Delta a \cdot t^2 = 0$, capped at 999 s when no positive root exists. For events in which the leader is moving away or stationary, time-to-collision and MTTC variables receive a sentinel value of 999 and are excluded from the clustering input to prevent the sentinel from distorting Euclidean and Gaussian assumptions.

#### 2.1.2. Family B — Profile Shape and Mechanism

Family B captures the shape or the evolvement of the deceleration profile and its asymmetry — the way the slowdown curve unfolds in time, and whether that curve is front-loaded (a sharper onset followed by a gentler release) or back-loaded (a gentler onset followed by a sharper release) — anchored in the profile-perspective literature [11,14,15]. The deceleration profile is resampled to 20 points on normalized event time, and a dataset-wide functional Principal Components Analysis is fitted, retaining two components (pca_shape1, pca_shape2). Asymmetry is summarized by ramp gradients computed between event onset, peak deceleration, and event end:

$$g_{up} = \frac{a_{peak} - a_0}{t_{peak} - t_0} \tag{3}$$

$$g_{down} = \frac{a_T - a_{peak}}{t_T - t_{peak}} \tag{4}$$

$$g_{asym} = |g_{up}| - |g_{down}| \tag{5}$$

where $a_0$ and $t_0$ are the longitudinal acceleration and time at event onset, $a_{peak}$ and $t_{peak}$ are values at peak deceleration, and $a_T$ and $t_T$ are values at event end. Positive $g_{asym}$ indicates a sharper onset than release; negative values indicate a gentler onset followed by a sharper release.

A mechanism proxy is computed as the ratio of peak absolute jerk to peak absolute deceleration:

$$m = \frac{|j_{peak}|}{|a_{peak}|} \tag{6}$$

Values approaching unity are characteristic of smooth coasting decelerations; values substantially above unity correspond to brake-pedal-like jerk applications. This proxy captures the coasting-versus-braking distinction established by Li et al. [15] and Da Lio et al. [14] without requiring pedal sensors.

#### 2.1.3. Family C — Scene Context

Family C captures the surrounding scene context of each event — what was happening around the vehicle, including the surrounding traffic, the presence and proximity of vulnerable road users, and the geometry of the road ahead. Continuous variables include the count and minimum distance of vulnerable road users (pedestrians, bicyclists, motorcyclists, wheeled riders, wheelchairs, strollers, wheeled devices) within 30 m of the follower at event onset; the count of regular vehicles within 50 m; the pair age, defined as the time elapsed since the current leader–follower pair was first observed; and the distance to the nearest stop sign, construction cone or barrel, pedestrian crossing, and intersection. The pair-age variable follows Fu et al. [36] and Hu et al. [37] as a proxy for the cut-in-versus-continuous-following distinction. Boolean variables include the lead vehicle class (heavy-vehicle indicator) and the vulnerable-road-user-on-path indicator, derived from a 2-second linear projection of every nearby vulnerable road user into the follower's longitudinal corridor (±2 m laterally, ≤30 m longitudinally), following Noonan et al. [43] and Chen et al. [44]. A derived categorical variable assigns each event to a coarse scene-type bucket (INTERSECTION_APPROACH, VRU_ADJACENT, CONSTRUCTION_ZONE, HEAVY_LEAD, DENSE_TRAFFIC, OPEN_ROAD) by applying a priority-ordered rule chain.

Together, the three families capture three complementary aspects of each event: how hard the deceleration was and how safety-critical the state at its onset (Family A); how it unfolded in time and whether it looked like coasting or braking (Family B); and what was happening in the surrounding scene at the moment it began (Family C). *Families A and B together form the 19-dimensional kinematic feature vector that drives the mode discovery; Family*

*C is held out from the clustering input so that the discovered modes are purely kinematic, and the strength of the kinematic-to-context association can be quantified separately*.

*2.2. Event Selection*

Events are extracted through a three-stage procedure: trajectory reconstruction, pair construction, and sustained-deceleration detection.

2.2.1. Trajectory Reconstruction

For each log, the per-frame ego-vehicle SE(3) pose is applied to each cuboid annotation to obtain city-frame coordinates. Per-track position sequences are smoothed by a Savitzky–Golay filter (window = 7 frames, polynomial order = 3), and longitudinal and lateral velocity, acceleration, and jerk are derived in the heading-aligned frame. Tracks shorter than ten frames are excluded, as are tracks with maximum longitudinal speed above 35 m/s. This stage retains 37,134 vehicle tracks from 3,605,320 agent-frame observations.

2.2.2. Pair Construction

Leader–follower pairs are identified frame-by-frame using a heading-aligned corridor: a candidate leader must lie within 50 m longitudinally ahead of the follower, within ±1.5 m laterally, and at a heading within 10° of the follower's heading. The 10° heading tolerance is relaxed from the 5° convention adopted by Yang et al. [17] to accommodate the broader scene-type composition of Argoverse 2, which includes turning maneuvers at intersections. Frames meeting the joint constraint are grouped into episodes, and episodes shorter than 1.0 s are discarded. This stage retains 4,365 stable pair episodes from 285 of 300 logs. The median pair-episode duration is 3.90 s, the median pair gap is 21.0 m, and the median time-headway is 3.56 s.

2.2.3. Sustained-Deceleration Detection

Within each pair episode, the follower's longitudinal acceleration sequence is scanned for intervals of sustained deceleration. A candidate event begins when the longitudinal acceleration $a_{long} \leq -0.5$ m/s² and ends when $a_{long} > -0.1$ m/s² (recovery threshold) or when the pair episode terminates. Candidate events shorter than 1.0 s and events truncated by pair-episode termination are discarded. The primary detection threshold of −0.5 m/s² inherits from the present author's prior work [63] and the urban convention adopted in [17]; robustness thresholds at −0.3 m/s² and −2.5 m/s² are noted for sensitivity analysis but are not the primary focus. This stage yields 1,219 deceleration events from 234 logs and 1,058 unique pairs.

*2.3. Analytical Methods*

2.3.1. Mode Discovery

The 19-dimensional kinematic vector (Families A and B, with TTC and MTTC excluded) is standardized by Z-score normalization. Three unsupervised clustering algorithms are fitted in parallel: HDBSCAN (min_cluster_size = 30, min_samples = 10); K-means at $K \in \{3, 4, 5, 6\}$ with 20 initialization and random seed 42; and Gaussian Mixture Models at $K \in \{3, 4, 5, 6\}$ with full covariance and five initializations. Algorithm choice is governed by an integrated criterion combining the silhouette score, the Davies–Bouldin index, the Bayesian Information Criterion, and bootstrap Adjusted Rand Index stability across 50 resamples. The Family C scene-context variables are deliberately held out from the clustering input so that the discovered taxonomy is purely kinematic and the context-modulation analysis remains unconfounded.

2.3.2. Context Modulation

The dependence of cluster membership on each Family C variable is tested with the appropriate statistical procedure. Categorical context variables (scene_type_coarse,

lead_is_heavy, vru_on_path) are tested via chi-square independence with Cramér's V as the effect size [65]:

$$V = \sqrt{\frac{x^2}{n \cdot (\min(r, c) - 1)}} \quad (7)$$

where $x^2$ is the chi-square statistic, n is the total sample size, and r and c are the number of rows and columns. Continuous context variables (VRU counts and distances, vehicle density, pair age, distances to static objects and high-definition map features) are tested via Kruskal–Wallis rank-sum [66] with epsilon-squared as the effect size [67]:

$$\varepsilon^2 = \frac{H - k + 1}{n - k} \quad (8)$$

where H is the Kruskal–Wallis statistic, k is the number of groups (clusters), and n is the total sample size. P-values are jointly corrected for multiple comparisons across the eleven tests using the Benjamini–Hochberg false discovery rate procedure [68]. Effect-size magnitudes are interpreted per conventional bands of 0.01–0.06 (small), 0.06–0.14 (medium), and ≥0.14 (large).

2.3.3. Speed-Regime Stratification

Events are stratified by velocity at event onset following the convention adopted by Qin et al. [62]: low-speed regime (LSR, $v < 5$ m/s), medium-speed regime (MSR, $5 \le v < 15$ m/s), high-speed regime (HSR, $v \ge 15$ m/s). K-means is re-fitted within each regime using the same kinematic feature set, with K capped at the number of dataset-wide clusters present in the regime. The Adjusted Rand Index is computed between the regime-specific labels and the dataset-wide labels restricted to that regime's events. A high Adjusted Rand Index (≥ 0.6) is interpreted as indicating regime-invariant modes; a low value as indicating regime-dependent modes.

2.3.4. Early-Event Classification

The classification target is the cluster label assigned by the primary algorithm, treated as a four-class problem in which all clusters — including the small outlier cluster — are retained. For each event and each of three prediction horizons $h \in \{0.5, 1.0, 1.5\}$ s, an early-window feature vector is extracted from the first h seconds of the event's follower trajectory; this vector includes early-window mean and minimum acceleration, early-window peak and root-mean-square jerk, early-window acceleration slope, and event-onset velocity, gap, Δv, gap-closing rate, looming, and DRAC. Three classifier variants are compared: M1 uses early kinematics only; M2 adds scene context; M3 uses scene context only as a lower-bound baseline. All variants use HistGradientBoostingClassifier with class_weight set to balanced. Cross-validation uses StratifiedGroupKFold with k = 5 on log_id, ensuring events from a given log never cross train–test boundaries. Permutation importance is computed at the primary horizon (1.0 s) on M2 using 20 repeats and macro-F1 as the scoring criterion.

## 3. Results

### *3.1. Event Set and Kinematic Distributions*

The full event set comprises 1,219 sustained deceleration events drawn from 234 of the 300 logs and from 1,058 unique leader–follower pairs. Event durations are right-skewed: the p10–p90 range is 1.20–4.50 s with a median of 2.20 s. Peak deceleration values cluster between mild and moderate magnitudes: 96 events (7.9%) reach the hard-deceleration regime ($a_{peak} \le -2.5$ m/s²), 649 events (53.2%) fall in the medium regime ($-2.5 < a_{peak} \le -1.0$), and 474 events (38.9%) sit in the mild regime ($-1.0 < a_{peak} \le -0.5$). The median peak deceleration is −1.19 m/s². Initial gap and time-headway distributions are consistent with urban car-following conventions: the median initial gap is 21.3 m and the median initial time-headway is 2.68 s. Notably, no event in the data has an initial time-to-collision ≤ 2 s, indicating that the Argoverse 2 urban driving population is dominated by anticipatory rather than emergency deceleration responses. The leader-class composition is REGULAR_VEHICLE-dominated at 91.6%, with the remaining 8.4% distributed across BOX_TRUCK, LARGE_VEHICLE, BUS, TRUCK, SCHOOL_BUS, ARTICULATED_BUS, and TRUCK_CAB categories.

The functional Principal Components Analysis on the resampled deceleration profile retains two components that capture 49.4% and 27.3% of profile-shape variance respectively, with cumulative variance 76.7%. This indicates that two principal modes of profile shape account for the substantial majority of profile variability.

*3.2. Mode Discovery*

The unsupervised mode-discovery procedure yields a single dominant algorithm choice. HDBSCAN labels every event as noise (zero clusters formed), indicating that the kinematic feature space has continuous gradient structure rather than density-separable modes. K-means produces stable solutions across $K \in \{3, 4, 5, 6\}$, with a clear peak in silhouette and bootstrap-ARI stability at K = 4: silhouette = 0.285, Davies–Bouldin = 1.158, and bootstrap Adjusted Rand Index = 0.897 ± 0.054 across 50 resamples. K = 6 shows substantial stability degradation (bootstrap ARI = 0.620 ± 0.202). Gaussian Mixture Models exhibit monotonically decreasing Bayesian Information Criterion across the K range (from −6,651 at K = 3 to −13,375 at K = 6), consistent with overfitting. **Table 1** summarises the algorithm comparison. K-means at K = 4 is selected as the primary algorithm.

The four discovered modes (**Figure 2**) have clean interpretations from their un-standardised centroid characteristics. Cluster 0 (n = 373, 30.6%) is characterized by strong gap-closing dynamics: median gap-closing rate of 3.34 m/s, median looming of 0.150 $s^{-1}$, median DRAC of 0.274 $m/s^2$, median Δv of −3.73 m/s. The cluster captures reactive closing events in which the follower is actively catching up to a slower or decelerating leader. Cluster 2 (n = 765, 62.8%) is the modal urban deceleration mode: median gap-closing rate of 0.72 m/s, median Δv of −1.11 m/s, low DRAC (0.015 $m/s^2$), mild peak deceleration (median −0.91 $m/s^2$) — anticipatory soft decelerations characteristic of non-emergency urban driving. Cluster 3 (n = 59, 4.8%) is distinguished by extreme jerk-magnitude characteristics (visible as the brake-like jerk in **Figure 3**): median peak jerk of 50.5 $m/s^3$ and median mechanism proxy of 23.1, with mild mean deceleration (median −0.79 $m/s^2$) and gentle ramp-up — the signature of brake-like jerk applications. Cluster 1 (n = 22, 1.8%) presents extreme outlier statistics on multiple variables simultaneously and is treated as a documented outlier cluster rather than as a behavioural mode.

Table 1. Algorithm comparison for unsupervised mode discovery. K-means K = 4 is selected as the primary algorithm.

| Algorithm | K / $n_{clu}$ | Silhouette | DB index | Boot-ARI / BIC | Noise % |
|---|---|---|---|---|---|
| HDBSCAN | 0 | — | — | — | 100.0% |
| K-means K=3 | 3 | 0.248 | 1.341 | 0.824 ± 0.203 | — |
| **K-means K=4 (primary)** | **4** | **0.285** | **1.158** | **0.897 ± 0.054** | **—** |
| K-means K=5 | 5 | 0.227 | 1.247 | 0.859 ± 0.080 | — |
| K-means K=6 | 6 | 0.218 | 1.306 | 0.620 ± 0.202 | — |
| GMM K=3 | 3 | 0.215 | — | BIC = −6,651 | — |
| GMM K=4 | 4 | 0.178 | — | BIC = −10,983 | — |
| GMM K=5 | 5 | 0.158 | — | BIC = −12,319 | — |
| GMM K=6 | 6 | 0.122 | — | BIC = −13,375 | — |

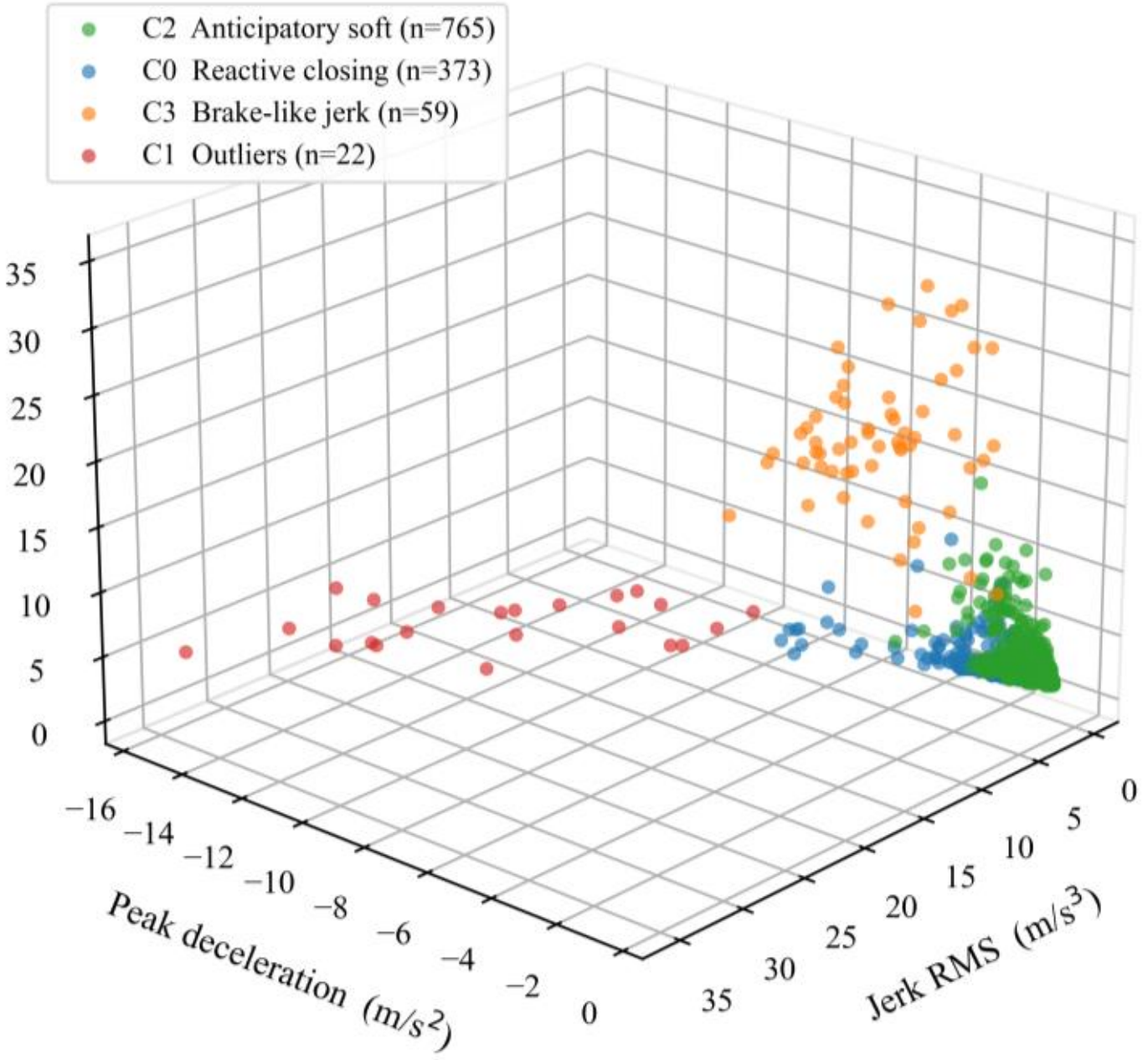

Figure 2. Each of the 1,219 events plotted in three discriminative dimensions: jerk RMS (m/s³), peak deceleration (m/s²), and the mechanism proxy. Color indicates the K-means K = 4 cluster assignment: C0 (reactive closing) blue, C1 (outliers) red, C2 (anticipatory soft) green, C3 (brake-like jerk) orange.

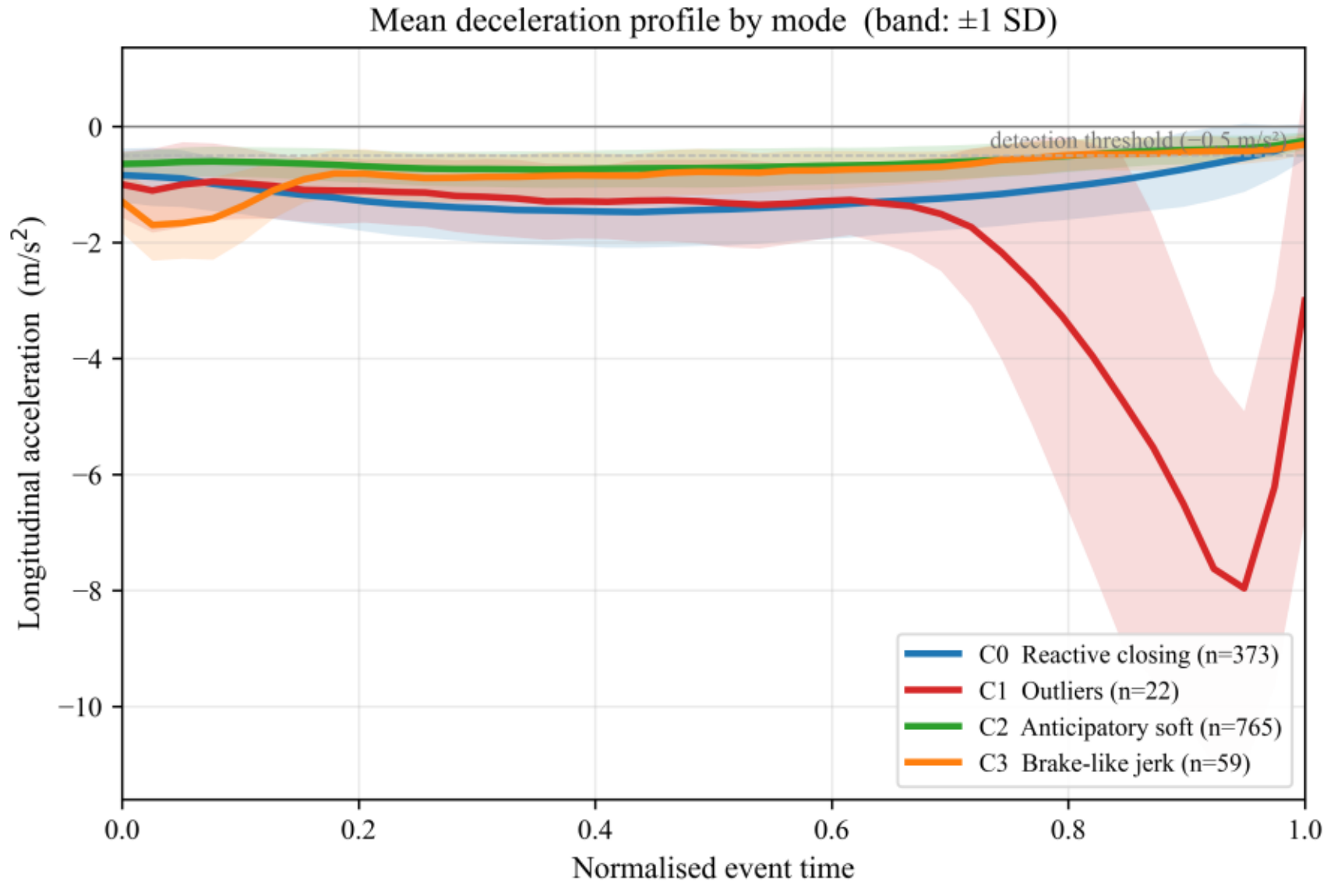

Figure 3. Mean longitudinal acceleration profile for each cluster on normalized event time, with shaded bands indicating ±1 SD. The detection threshold (−0.5 m/s²) is shown as a dashed reference line.

### 3.3. Context Modulation

Cross-tabulation of the primary cluster label against each of the eleven scene-context variables reveals one variable with a medium effect and a tail of small-to-negligible effects. The Kruskal–Wallis test on pair age produces H = 106.52, $p = 6.15 \times 10^{-23}$, $p_{BH} = 6.77 \times 10^{-22}$, $\varepsilon^2 = 0.085$ — the only context variable to meet the medium-effect threshold. Per-cluster medians of pair age stratify cleanly (**Figure 4**): 1.0 s for Cluster 0, 1.4 s for Cluster 1, 2.8 s for Cluster 2, 4.7 s for Cluster 3. The 2.8-fold span between Clusters 0 and 2, and the additional 1.7-fold span between Clusters 2 and 3, supports the cut-in-versus-established-follower interpretation introduced by Fu et al. [36] and Hu et al. [37].

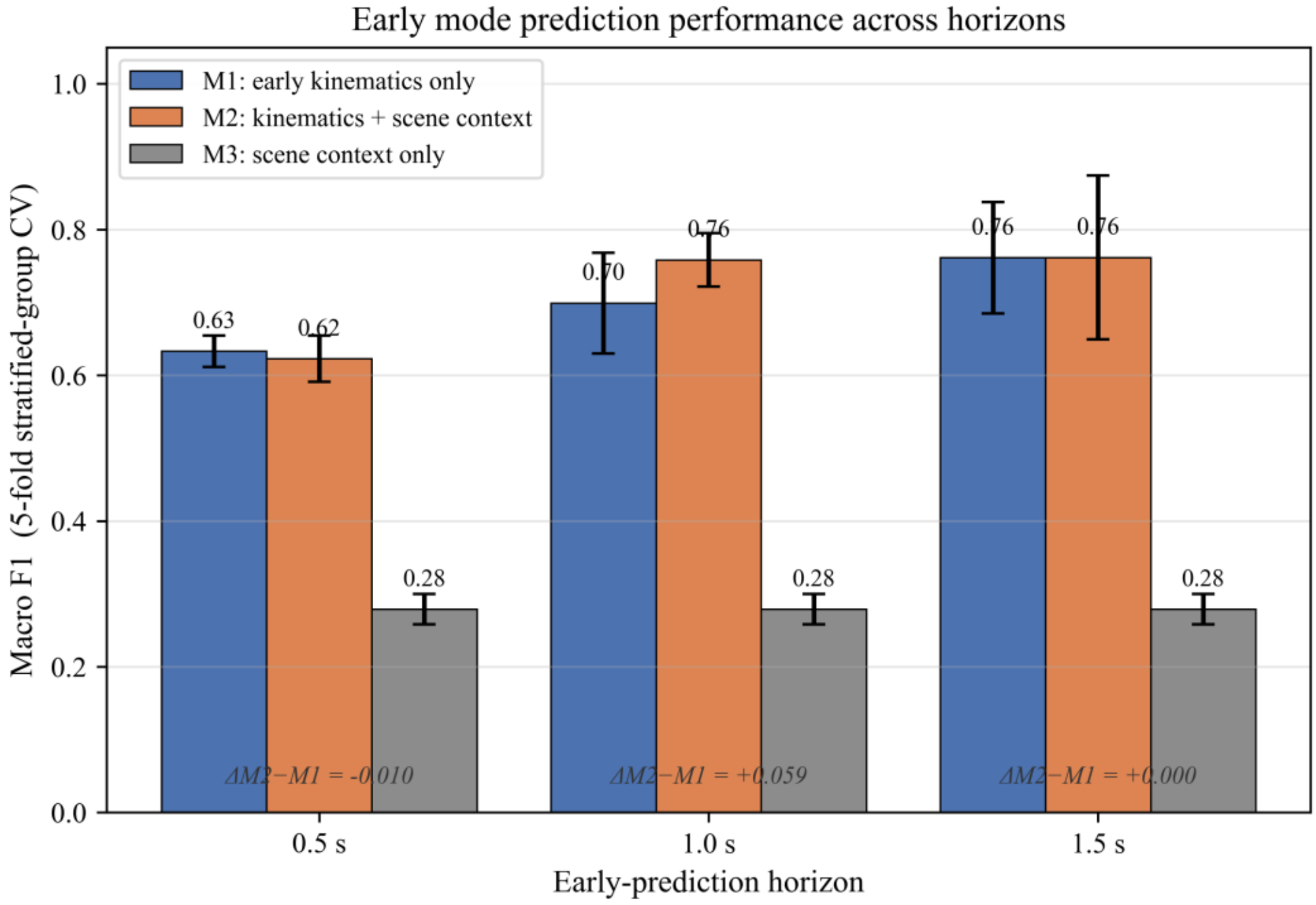

Figure 4. Three-dimensional kernel-density ridges of pair age (seconds since pair formation) for each cluster. Median pair age per cluster is annotated in the legend. Cluster 0 shows the youngest pairs, consistent with cut-in-induced decelerations; Cluster 3 shows the oldest established pairs.

All scene-geometry variables show smaller effects. The categorical scene_type_coarse test does not reach significance after Benjamini–Hochberg correction ($\chi^2$ = 15.27, $p_{BH}$ = 0.475, V = 0.065). The Kruskal–Wallis tests on vehicle density ($\varepsilon^2$ = 0.014), distance to construction barrels ($\varepsilon^2$ = 0.017), and distance to stop sign ($\varepsilon^2$ = 0.012) yield significant raw p-values but small effect sizes. Distances to crosswalks and intersections show essentially negligible effects ($\varepsilon^2 \leq 0.001$). VRU presence indicators yield small effects (V or $\varepsilon^2 \leq 0.073$). The lead_is_heavy indicator is essentially flat across clusters (V = 0.024, $p_{BH}$= 0.875). The combined interpretation: the four modes are pair-age-modulated but otherwise largely independent of scene context.

### *3.4. Speed-Regime Stratification*

Stratification by velocity at event onset reveals strong regime dependence at low speed and regime invariance at medium speed. The low-speed regime (LSR, $v_{start}$< 5 m/s) contains 198 events distributed across the dataset-wide clusters as {C0: 30, C1: 1, C2: 167} — Cluster 3 is essentially absent. Within-regime re-clustering yields a regime-specific partition with K = 3 and cluster sizes {0: 87, 1: 104, 2: 7} that does not align with the dataset-wide partition: the Adjusted Rand Index between the two label assignments is 0.166, indicating regime-dependent modes. The medium-speed regime (MSR, 5 ≤ $v_{start}$ < 15 m/s) contains 999 events, accounting for 82.0% of the data, and exhibits all four dataset-wide clusters with sizes proportional to the overall distribution. Regime-specific re-clustering at K = 4 yields a partition with ARI = 0.817 against the dataset-wide labels, indicating regime-invariant modes. The high-speed regime (HSR, $v_{start}$≥ 15 m/s) contains only 22 events, distributed as {C0: 8, C2: 13, C3: 1}; this sample is insufficient for within-regime re-clustering.

### *3.5. Early-Event Classification*

Mode-classification results at three observation horizons are summarized in Table 2. With M1 representing kinematics only, M2 representing kinematics + scene context and M3 representing scene context only.

Table 2. Mode-classification performance at three early-prediction horizons and three model variants. M1 uses early kinematics only; M2 adds scene context; M3 uses scene context only. Macro and weighted F1 reported as mean ± SD across five stratified-group folds on log_id.

| Horizon | Variant | Macro F1 | Weighted F1 | C0 F1 | C1 F1 | C2 F1 | C3 F1 |
|---|---|---|---|---|---|---|---|
| 0.5 s | M1 | 0.633 ± 0.022 | 0.823 ± 0.016 | 0.74 | 0.00 | 0.88 | 0.91 |
| 0.5 s | M2 | 0.623 ± 0.032 | 0.825 ± 0.026 | 0.75 | 0.00 | 0.88 | 0.86 |
| 0.5 s | M3 | 0.279 ± 0.021 | 0.555 ± 0.033 | 0.37 | 0.00 | 0.70 | 0.04 |
| 1.0 s | M1 | 0.699 ± 0.069 | 0.854 ± 0.013 | 0.79 | 0.25 | 0.90 | 0.85 |
| **1.0 s** | **M2** | **0.758 ± 0.037** | **0.861 ± 0.024** | **0.80** | **0.46** | **0.90** | **0.87** |
| 1.0 s | M3 | 0.279 ± 0.021 | 0.555 ± 0.033 | 0.37 | 0.00 | 0.70 | 0.04 |
| 1.5 s | M1 | 0.761 ± 0.076 | 0.868 ± 0.012 | 0.81 | 0.48 | 0.91 | 0.85 |
| 1.5 s | M2 | 0.762 ± 0.113 | 0.869 ± 0.028 | 0.82 | 0.51 | 0.91 | 0.81 |
| 1.5 s | M3 | 0.279 ± 0.021 | 0.555 ± 0.033 | 0.37 | 0.00 | 0.70 | 0.04 |

Pair-age distribution by mode (cut-in vs. established follower)

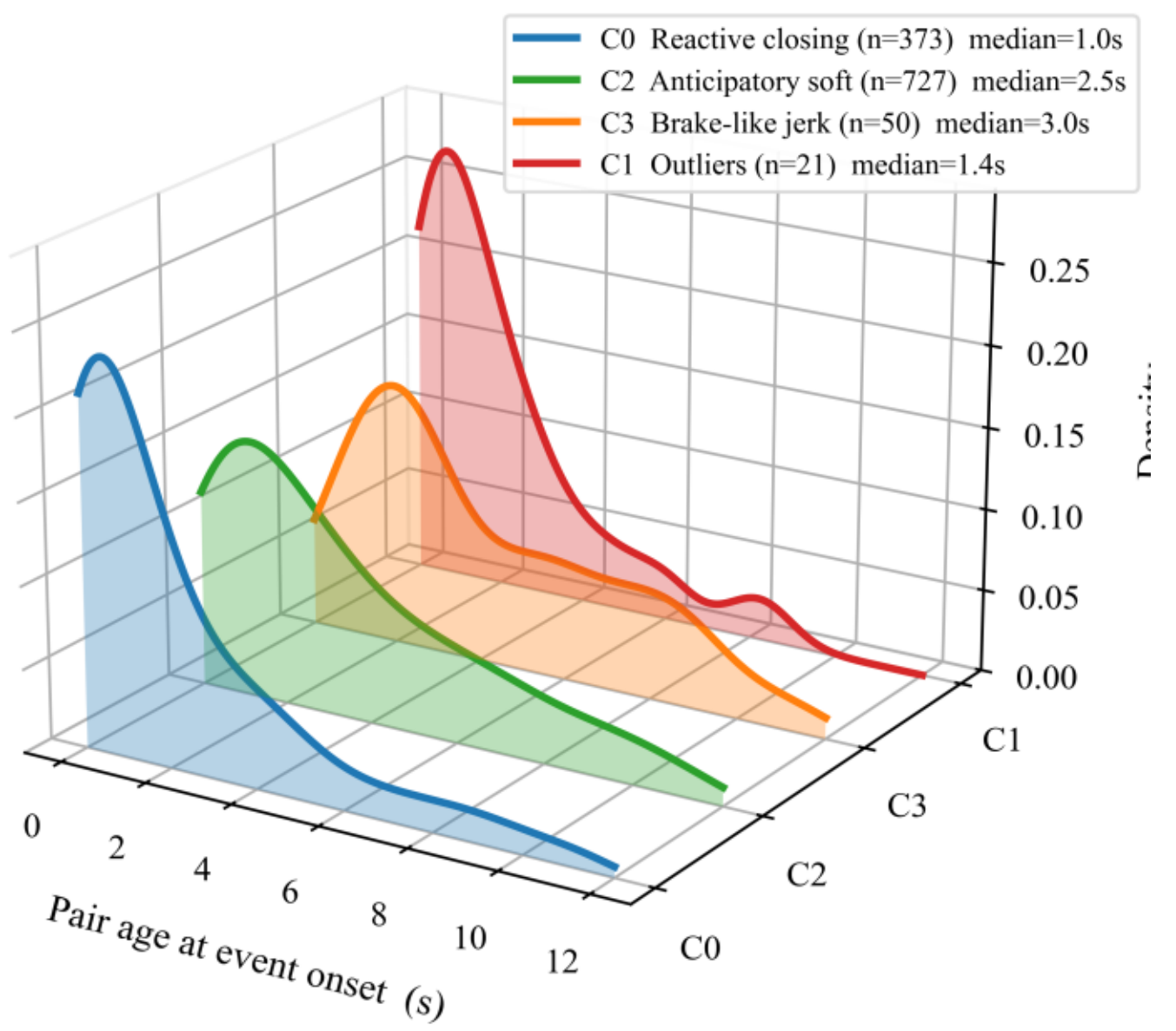

Figure 5. Macro F1 by prediction horizon and classifier variant. Bars are mean ± SD across five stratified-group folds on log_id. ΔM2−M1 annotated at each horizon quantifies the marginal value of adding scene context: negative at 0.5 s, positive at 1.0 s, zero at 1.5 s.

The headline finding is at h = 1.0 s with M2 (kinematic and context combined; **Figure 5**): macro-F1 = 0.758 ± 0.037 across five stratified-group folds, with weighted-F1 = 0.861 ± 0.024. Per-class F1 values at this configuration: 0.80 (Cluster 0), 0.46 (Cluster 1), 0.90 (Cluster 2), 0.87 (Cluster 3). The substantially lower F1 for Cluster 1 reflects its small sample (n = 22); per-class precision for Cluster 1 remains high (0.87) but recall is 0.32.

The marginal value of scene context (M2 minus M1 macro-F1) varies systematically with horizon: −0.010 at h = 0.5 s, +0.059 at h = 1.0 s, +0.000 at h = 1.5 s. At the shortest horizon, the addition of scene context slightly degrades performance, suggesting that the early-kinematic signal is so noisy that context features act as distractors. At the primary horizon, scene context contributes a meaningful 6 macro-F1 points. At the longest horizon, scene context becomes redundant. The context-only baseline (M3) yields macro-F1 = 0.279 independently of horizon, only marginally above the random-guess baseline for a four-class problem with skewed support.

Permutation importance on M2 at h = 1.0 s **(Figure 6**) identifies early-window jerk as the dominant predictive feature: j_rms_early contributes Δ macro-F1 = 0.211 ± 0.037 when shuffled, and j_peak_early contributes Δ macro-F1 = 0.094 ± 0.030. Together these two features account for substantially more importance than the next thirteen features combined. The remaining top-15 ranks are dominated by additional kinematic features (a_mean_early, drac_at_start, dv_at_start, a_slope_early, gap_close_rate_at_start, a_min_early, looming_at_start, v_long_at_start), with a small number of context features (vru_min_dist_m, pair_age_s, dist_intersection_m, dist_crosswalk_m, scene_INTERSECTION_APPROACH) appearing at the bottom with importance values an order of magnitude below the kinematic leaders.

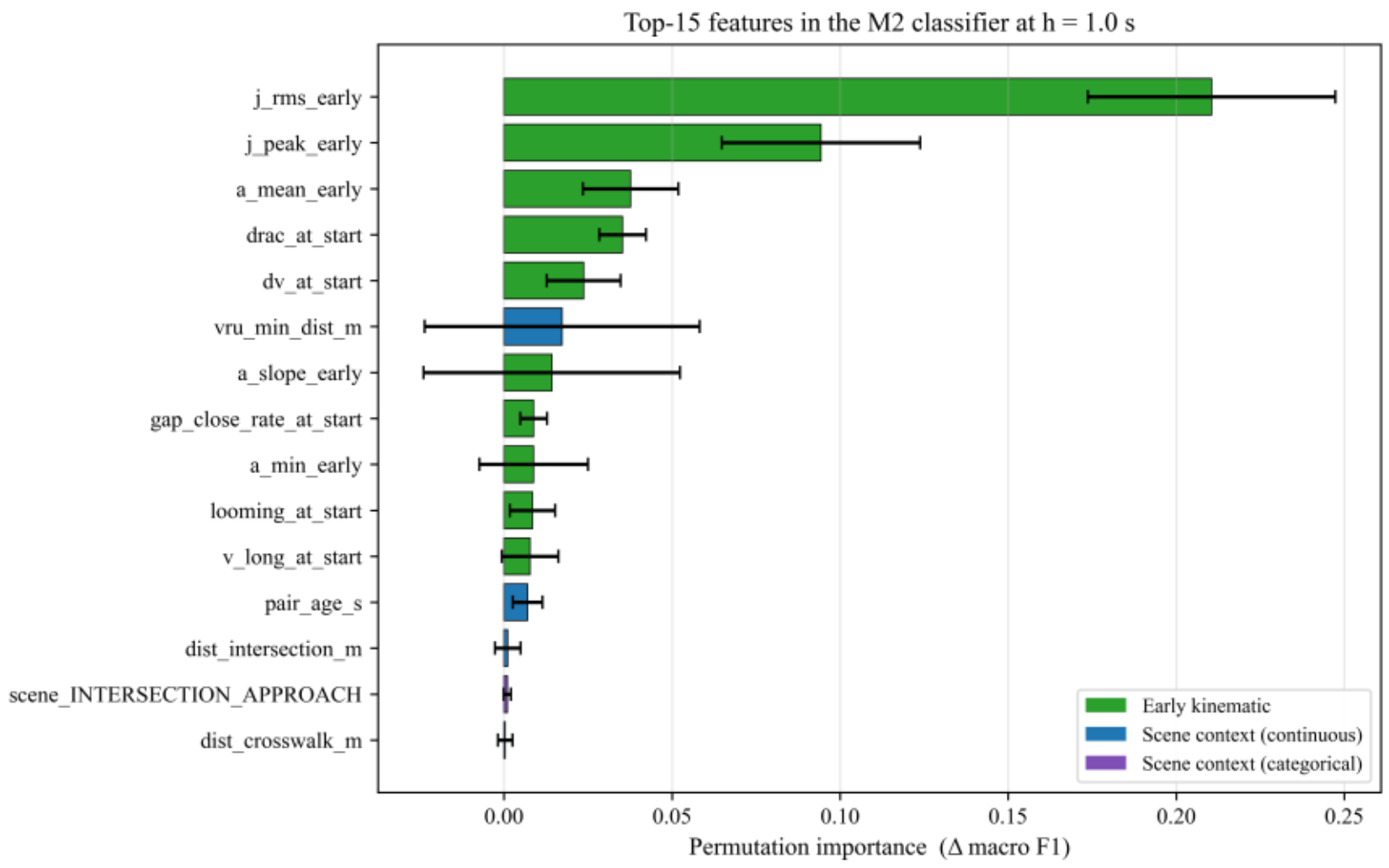

Figure 6. Top-15 permutation importances for the M2 classifier at h = 1.0 s. Bars are coloured by feature kind: green for early-window kinematics, blue for continuous scene context, purple for categorical scene context. Error bars show standard deviation across 20 permutation repeats.

The confusion matrix at h = 1.0 s, M2 **(Figure 7**) shows clean diagonals for the three large clusters (C0: 287/373 = 76.9% recall; C2: 705/765 = 92.2%; C3: 54/59 = 91.5%) and confirms the recall limitation on the small Cluster 1 (7/22 = 31.8% recall).

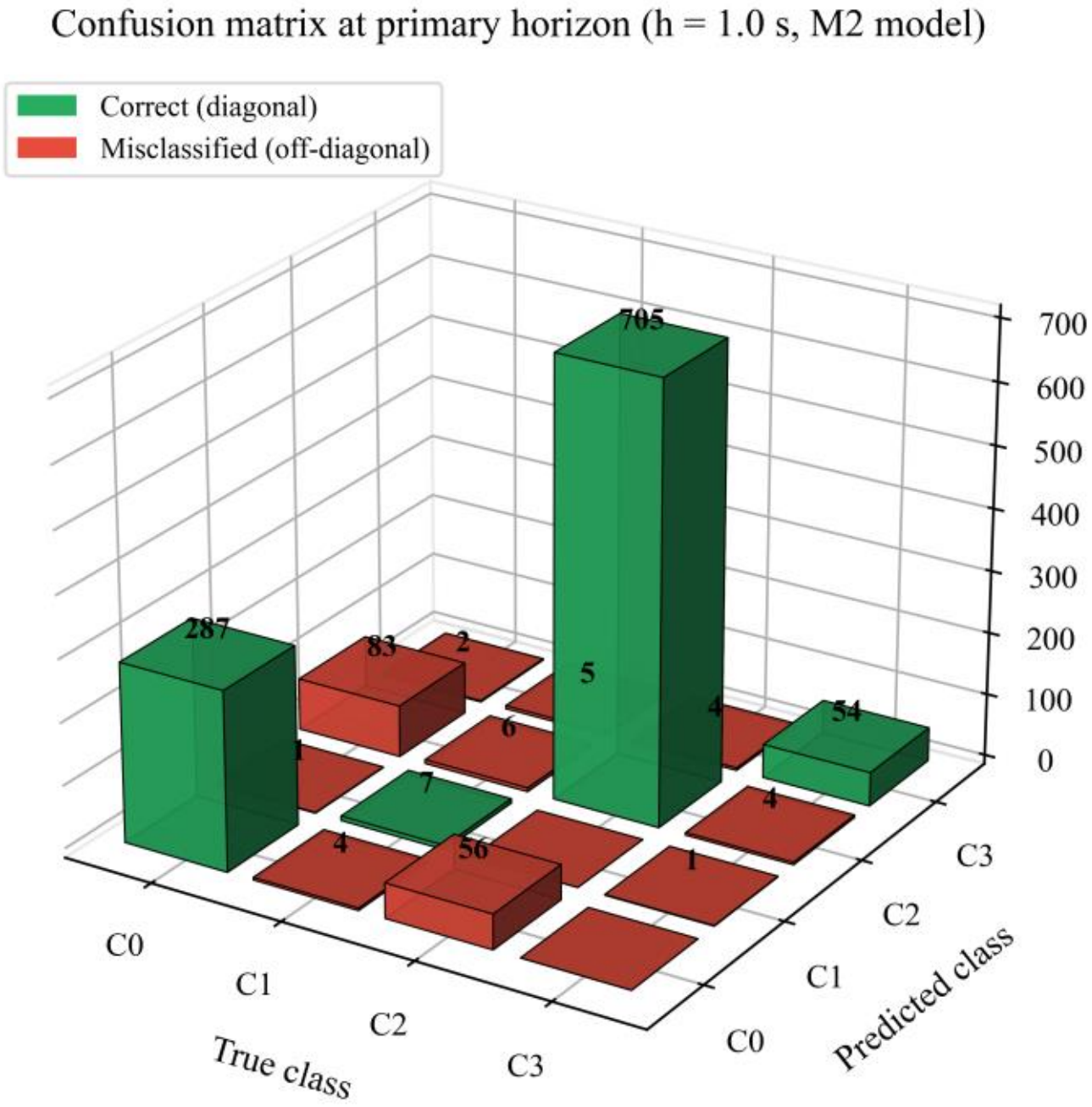

Figure 7. Confusion matrix at h = 1.0 s, M2 rendered as 3-D bars. Diagonal bars (green) show correct predictions; off-diagonal bars (red) show misclassifications. The dominant off-diagonal cell is C0→C2 (83 events), indicating that the most common error mode is reactive-closing events misclassified as anticipatory-soft.

## 4. Discussion

### *4.1. Main Findings*

The application of the assembled pipeline to 1,219 urban deceleration events yields five empirical findings, each extending a specific concept from the threads reviewed in Section 1.1 by quantifying it for the urban perception-AV regime.

First, the $K = 4$ partition discovered in the present data aligns with the three-to-four-cluster convergence repeatedly documented in the unsupervised driving-style literature [20,22,24–27,35]. Bootstrap Adjusted Rand Index stability of 0.897 indicates that the taxonomy is reproducible rather than method-dependent. The complete failure of HDBSCAN to identify density-separable clusters is itself informative: the urban deceleration data has continuous gradient structure rather than crisp density modes.

Second, the partition between Cluster 2 ($n = 765$; median mechanism proxy 2.6, median peak jerk 2.3 m/s$^3$) and Cluster 3 ($n = 59$; median mechanism proxy 23.1, median peak jerk 50.5 m/s$^3$) reproduces the coasting-versus-braking distinction documented by Li et al. [15] with pedal-instrumented vehicles and the discrete stop-phase partition of Da Lio et al. [14]. Recovering this distinction from trajectory data alone supports the mechanism-proxy feature as transferable for behavioural-mechanism inference in trajectory-only datasets, of relevance to the FollowNet benchmark [7].

Third, Cluster 0 ($n = 373$) is sharply characterized by gap-closing dynamics (median gap-closing rate 3.34 m/s, median looming 0.150 s$^{-1}$), reproducing the looming-driven response of Xue et al. [1] and demonstrating that the looming-anchored deceleration is a separable behavioural mode. This is the modal urban analogue of the highway looming-driven brake response documented in [63].

Fourth, among eleven scene-context variables, only pair age shows a medium effect ($\varepsilon^2 = 0.085$) on mode membership. The cleanly stratified per-cluster medians — 1.0, 1.4, 2.8, and 4.7 s — capture the cut-in-versus-established-follower distinction documented by Fu et al. [36], Hu et al. [37], Zhao et al. [38], and Gu et al. [39]. Scene-geometry and VRU presence indicators show negligible-to-small effects ($\varepsilon^2$ or $V \leq 0.073$) despite high statistical power (N = 1,219). The four discovered modes are therefore pair-age-modulated but otherwise largely independent of scene context.

Fifth, the macro-F1 of M2 (kinematic + context) at h = 1.0 s reaches 0.758, with kinematic-only M1 at 0.699 and context-only M3 at 0.279. The marginal value of context varies systematically with horizon: slightly negative at 0.5 s, positive at 1.0 s (+0.059), zero at 1.5 s. This extends the early-prediction tradition [2,3,36] by demonstrating that scene context is a horizon-conditional predictor. Permutation importance further establishes that early-window jerk dominates predictive contribution.

Beyond these five findings, the modes are regime-invariant in medium-speed urban driving (ARI = 0.817) and regime-dependent in low-speed driving (ARI = 0.166). Low-speed urban driving appears to follow a distinct set of deceleration patterns not captured by the dataset-wide medium-speed-dominated taxonomy, consistent with Qin et al. [62].

### *4.2. Recommendations for Future Research*

First, multi-dataset validation: replicating the taxonomy on the Argoverse 2 Motion Forecasting split and on other perception-AV datasets would test transferability, complementing the unified evaluation framework of [7]. Second, coverage of the high-speed regime: the present HSR cohort of 22 events blocks any HSR-specific claim; targeted sampling would enable a complete regime-stratified analysis along the lines of Liu and Zhang [12]. Third, threshold sensitivity sweeps at $-0.3$ m/s$^2$ [15] and $-2.5$ m/s$^2$ [18] would quantify robustness of the mode taxonomy. Fourth, integration with motion-planning frameworks: the early-mode classifier can be embedded as a behavioural-state estimator in model predictive control or behavioural-imitation pipelines, complementing Ma et al. [50] and the multi-leader recurrent model of Das et al. [8]. Fifth, mechanism-proxy validation against pedal-sensor-instrumented data would confirm or refine the coasting-versus-braking interpretation in dialogue with Li et al. [15]. Sixth, explanatory layering: coupling the mode taxonomy with the surrogate-safety-measure representation of Lu et al. [61], the field-theoretic frameworks of Joo et al. [51] and Liu and Xiang [52], and the SHAP-attributed aggression-recognition approach of Cheng et al. [47] would deepen interpretability.

### *4.3. Recommendations for Practice*

Practical implications for ADAS evaluation follow from the horizon-dependent classifier results. The 1.0 s observation horizon emerges as the operational sweet spot: kinematic-only classification reaches macro-F1 = 0.699 and the addition of scene context raises performance to 0.758. Aggressive 0.5 s prediction targets sacrifice the context contribution; 1.5 s prediction targets do not gain meaningfully over 1.0 s. Sensor-light ADAS deployments can rely on longitudinal kinematics alone for mode inference at 1.0 s with macro-F1 of approximately 0.70, accepting a six-percentage-point F1 penalty for omitting scene perception.

Low-speed urban driving requires separately calibrated behavioural models. The regime-dependence finding (ARI = 0.166 for LSR) indicates that a medium-speed-trained controller does not transfer cleanly to creep, stop-and-go, or queue-discharge scenarios. ADAS systems intended for operation across the LSR–MSR transition should explicitly model the speed-regime boundary, in line with Qin et al. [62]. ADAS pair-tracking infrastructure should preserve and surface the pair-age signal as a behavioural-context feature, given its medium effect size; the pair-establishment timestamp should be propagated through the perception–prediction–planning stack and exposed to downstream behavioural-mode classifiers, complementing Fu et al. [36], Hu et al. [37], Zhao et al. [38], and Gu et al. [42]. Surfacing of driver-style or preference features as proposed by Lyu et al. [9] could further disambiguate kinematically similar but contextually different events. Application of mode-aware ADAS reasoning to off-ramp and transition-area scenarios, where the cooperative adaptive cruise control results of Dong et al. [65] demonstrate measurable safety and capacity benefits, represents a concrete deployment context.

*4.4. Limitations*

The study has four substantive limitations. First, Cluster 1 contains only 22 events, and its per-class classification metrics (F1 = 0.46, recall = 0.32 at h = 1.0 s) are correspondingly noisy; the cluster is reported as a documented outlier category rather than as a fifth behavioural mode. Second, the high-speed regime contains only 22 events, too small for within-regime re-clustering, and no HSR-specific claim is offered. Third, the mechanism proxy captures the coasting-versus-braking distinction without direct pedal validation; although its interpretation is supported by alignment with Li et al. [15] and Da Lio et al. [14], direct cross-validation with pedal sensors is outside the scope of the present data. Fourth, the Argoverse 2 dataset contains no event with initial time-to-collision $\leq 2$ s, indicating that the present driving population is dominated by anticipatory rather than emergency deceleration; conclusions about emergency-deceleration behaviour are therefore outside the study's scope and require complementary datasets such as those characterized by Yamagishi [18] and Wang et al. [16].

## 5. Conclusions

This study assembled established conceptual and analytical tools from the deceleration, car-following, clustering, and interpretability literatures into a single coherent pipeline, and applied that pipeline to 1,219 sustained urban deceleration events drawn from the Argoverse 2 Sensor dataset [4,64]. Four stable behavioural modes — anticipatory soft, reactive closing, brake-like jerk, and an outlier category — emerged from K-means clustering on a 19-dimensional kinematic feature vector with bootstrap ARI stability of 0.897. Among eleven scene-context variables, only pair age modulated mode membership at a medium effect size ($\varepsilon^2 = 0.085$); scene geometry, VRU proximity, and intersection distance produced negligible-to-small effects. A HistGradientBoosting classifier predicted mode membership from the first 1.0 s of each event at macro-F1 = 0.758, with early-window jerk dominating predictive importance and scene context contributing +0.059 F1 over kinematics alone. The modes were regime-invariant in medium-speed driving but regime-dependent at low speed. By assembling established literature tools into a transferable methodology and applying that methodology to a perception-AV dataset for the first time in this configuration, the study extends deceleration-mode discovery from the earlier NGSIM-based highway regime into the urban perception-AV regime, with downstream applications to ADAS evaluation, behavioural-imitation modelling, and surrogate-safety-measure interpretation.

**Author Contributions:** Conceptualization, E.S.L.; methodology, E.S.L.; software, E.S.L.; validation, E.S.L.; formal analysis, E.S.L.; investigation, E.S.L.; resources, E.S.L.; data curation, E.S.L.; writing—original draft preparation, E.S.L.; writing—review and editing, E.S.L.; visualization, E.S.L. The author has read and agreed to the published version of the manuscript.

**Funding:** This research received no external funding.

**Institutional Review Board Statement:** Not applicable.

**Informed Consent Statement:** Not applicable.

**Data Availability Statement:** The Argoverse 2 Sensor Dataset analyzed in this study is publicly available at https://www.argoverse.org/av2.html. Derived event-level data, code, and analysis pipeline can be made available from the corresponding author upon reasonable request.

**Acknowledgments:** During the preparation of this manuscript the author used Claude Opus 4.7 Extra for the purpose of proper tense usage and grammar corrections. The author has reviewed and edited the output and takes full responsibility for the content of this publication.

**Conflicts of Interest:** The author declares no conflicts of interest.